\documentclass{article}
\pdfoutput=1

\usepackage[utf8]{inputenc} %
\usepackage[T1]{fontenc}    %
\usepackage{hyperref}       %
\usepackage{url}            %
\usepackage{booktabs}       %
\usepackage{amsfonts}       %
\usepackage{nicefrac}       %
\usepackage{microtype}      %
\usepackage{lipsum}
\usepackage{fancyhdr}       %
\usepackage{graphicx}       %
\usepackage{amsmath}
\usepackage{amssymb}        %
\usepackage{bm}
\usepackage{subcaption}
\usepackage[colorinlistoftodos]{todonotes}
\usepackage{algorithm, algorithmic}
\usepackage{subcaption}   %
\usepackage{svg}

\usepackage{hyperref}

\usepackage[accepted]{icml2023}

\graphicspath{{figures/}}

\usepackage[acronym,nowarn,hyperfirst=false]{glossaries}

\newacronym{PDE}{PDE}{partial differential equation}
\newacronym{CFD}{CFD}{computational fluid dynamics}
\newacronym[longplural={degrees of freedom}]
           {DOF}{DOF}{degree of freedom}
\newacronym{MSE}{MSE}{mean-squared error}
\newacronym{RMSE}{RMSE}{root-mean-square error}
\newacronym{ML}{ML}{machine learning}
\newacronym{NN}{NN}{neural network}
\newacronym{ANN}{ANN}{artificial neural network}
\newacronym{FNN}{FNN}{feedforward neural network}
\newacronym{CPWA}{CPWA}{continuous piecewise affine}
\newacronym{MLP}{MLP}{multilayer perceptron}
\newacronym{ReLU}{ReLU}{rectified linear unit}
\newacronym{SDF}{SDF}{signed distance function}
\newacronym{AM}{AM}{analytical marching}
\newacronym{HAM}{HAM}{hierarchical analytical marching}

\glsdisablehyper

\usepackage{mathtools}
\newcommand\V[1]{\mathbf{#1}} %
\AtBeginDocument{} %

\newcommand\R{\mathbb{R}} %

\DeclarePairedDelimiter\paren{\lparen}{\rparen}

\DeclareMathOperator*{\ReLU}{ReLU}

\def\x{\V{x}} %
\def\b{\V{b}} %
\def\W{\V{W}} %
\def\w{\V{w}} %

\icmltitlerunning{Polyhedral Complex Extraction from ReLU Networks using Edge Subdivision}

\begin{document}

\twocolumn[

\icmltitle{Polyhedral Complex Extraction from ReLU Networks using Edge Subdivision} %

\icmlsetsymbol{equal}{*}

\begin{icmlauthorlist}
\icmlauthor{Arturs Berzins}{SINTEF,UiO}
\end{icmlauthorlist}

\icmlaffiliation{SINTEF}{SINTEF, Oslo, Norway}
\icmlaffiliation{UiO}{Department of Mathematics, University of Oslo, Oslo, Norway}

\icmlcorrespondingauthor{Arturs Berzins}{arturs.berzins@sintef.no}

\icmlkeywords{ReLU, piecewise-linear, cellular-complex, edge, skeleton, subdivision}

\vskip 0.3in
]

\printAffiliationsAndNotice{}  %

\begin{abstract}
A neural network consisting of piecewise affine building blocks, such as fully-connected layers and ReLU activations, is itself a piecewise affine function supported on a polyhedral complex.
This complex has been previously studied to characterize theoretical properties of neural networks, but, in practice, extracting it remains a challenge due to its high combinatorial complexity.
A natural idea described in previous works is to subdivide the regions via intersections with hyperplanes induced by each neuron.
However, we argue that this view leads to computational redundancy.
Instead of regions, we propose to subdivide edges, leading to a novel method for polyhedral complex extraction.
A key to this are sign-vectors, which encode the combinatorial structure of the complex.
Our approach allows to use standard tensor operations on a GPU, taking seconds for millions of cells on a consumer grade machine.
Motivated by the growing interest in neural shape representation, we use the speed and differentiablility of our method to optimize geometric properties of the complex.
The code is available on GitHub\footnote{\href{https://github.com/arturs-berzins/relu_edge_subdivision}{github.com/arturs-berzins/relu\_edge\_subdivision}}.
\end{abstract}

\begin{figure}
\begin{center}
  \includegraphics[width=0.75\columnwidth]{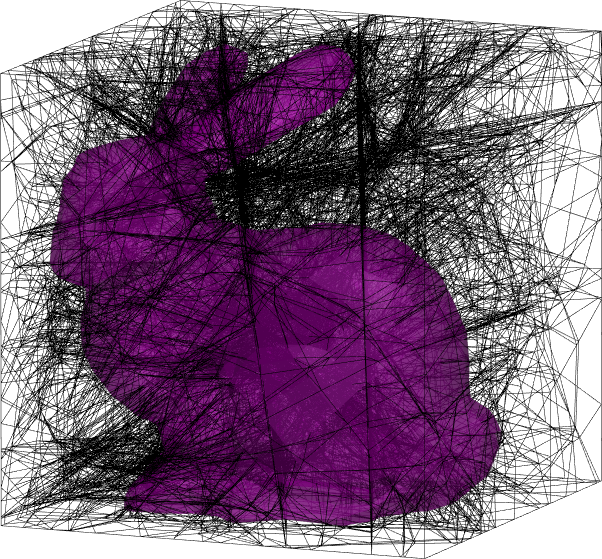}
\end{center}
\caption{Example of the polyhedral complex extracted from a ReLU NN trained on the signed distance field of a $D=3$ Stanford bunny.}
\label{fig:bunny}
\end{figure}

\begin{figure*}
    \centering
    \begin{subfigure}[b]{0.175\textwidth}
        \centering
        \includegraphics[width=\textwidth]{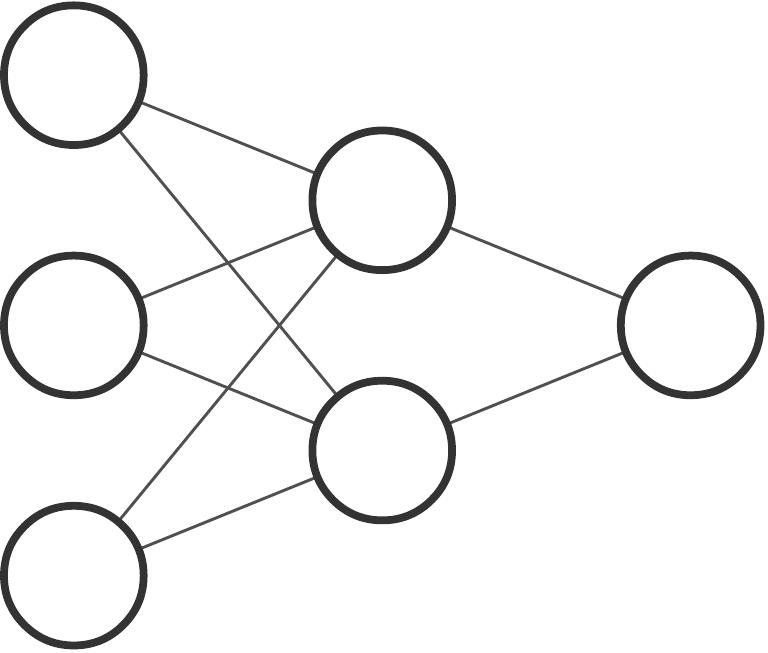}
        \label{fig:a}
    \end{subfigure}
    \hspace{2.5em}
    \begin{subfigure}[b]{0.65\textwidth}
        \centering
        \includegraphics[width=\textwidth]{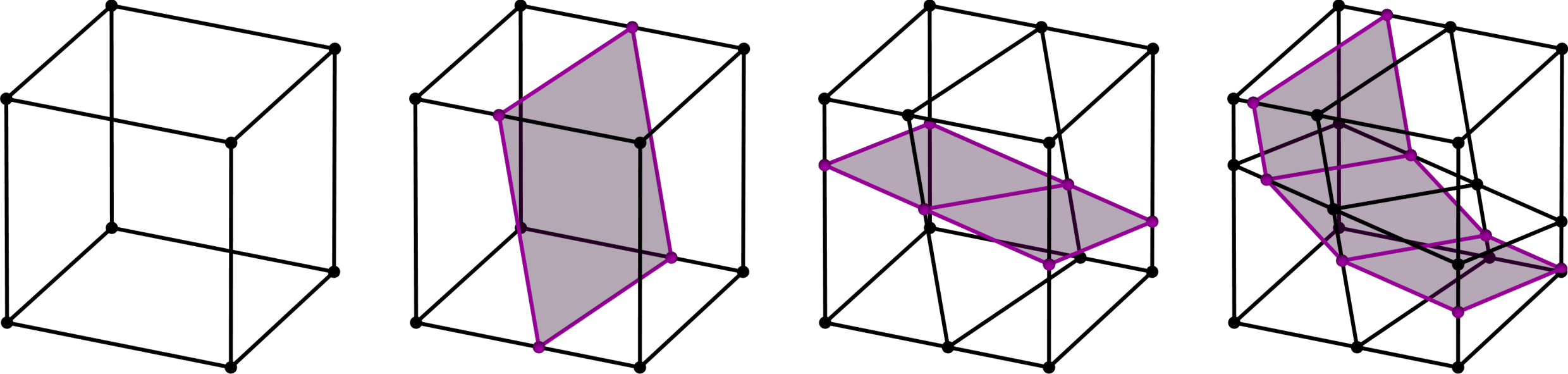}
        \label{fig:b}
    \end{subfigure}
    \vspace{-1.0\baselineskip}
    \caption{
    In \gls{CPWA} \glspl{NN}, each neuron of each layer sequentially subdivides the polyhedral complex.
    Each neuron of the first hidden layer contributes an affine hyperplane.
    Each neuron of the deeper layers contributes a folded hyperplane.
    Illustrated is the subdivision of a cubic domain in the $D=3$ input space by the shown \gls{NN}.
    While previous methods subdivide the regions (highest dimensional cells), our method subdivides edges.}
    \label{fig:ab}
\end{figure*}

\section{Introduction}
\label{sec:introduction}

A \gls{NN} is a \gls{CPWA} function if it is a composition of \gls{CPWA} operators, most notably fully-connected layers and \gls{ReLU} activations.
The \gls{CPWA} nature induces a discrete partitioning of the input domain, which provides an additional avenue to study \glspl{NN} in terms of their expressivity, robustness, training techniques, and unique geometry.

It is known that each affine piece is supported on a convex polyhedral set.
The collection of these polyhedral sets forms a polyhedral complex, which is induced by an arrangement of folded hyperplanes representing decision boundaries of neurons \cite{Grigsby22BentHyperplane}.
This complex has been linked to max-affine spline operators, which possess intriguing geometric properties and allow for joint optimization of the domain partition and spline coefficients \cite{Balestriero2021}.
\gls{CPWA} \glspl{NN} have also been used in neural implicit shape representations for boundary mesh extraction \cite{Lei2020} and visualization \cite{humayun2022exact, humayun2023splinecam}.

The diverse range of applications motivates a computational method to extract the complex.
A trivial approach is to evaluate neuron states at sampled points.
While this suffices to visualize an estimated domain partition or compute a lower bound on the number of regions, it does not provide the exact complex.
Proposed exact approaches include formulating this as a mixed-integer linear program \cite{Serra2018} or employing state-flipping \cite{Lei2020}.
However, the most natural and widely discussed approach is region subdivision, where the complex's regions are successively subdivided from neuron to neuron and layer to layer \cite{Raghu17, Hanin2019, Wang22estimation, humayun2022exact, humayun2023splinecam}.

Our method is motivated by the observation of redundancy in region subdivision.
The continuity of the activation function ensures that a folded hyperplane remains continuous across another fold (see Figure \ref{fig:continuous}).
However, considering each region independently leads to computing the same new vertex or, alternatively, identifying the same redundant hyperplane on all $2^{D-1}$ regions sharing a common edge, where $D$ is the dimension of the input space.
Our method alleviates this redundancy by leveraging continuity and disregarding the regions, instead using solely the unique vertices and edges, i.e. the 1-skeleton.
The key idea to \emph{edge subdivision} is to sequentially consider each neuron, i.e. folded hyperplane, evaluate all vertices with the \gls{NN} and compare the signs of a vertex pair sharing an edge.
If the signs differ, linear interpolation determines the location of a new vertex.
The edges containing the connectivity information are updated accordingly.
For this, we propose to leverage sign-vectors, which indicate the pre-activation sign of every neuron at every point or for every cell of the complex and altogether encode the combinatorial structure of the whole complex.

Our edge subdivision approach is naturally parallel and the use of sign-vectors affords additional structure, which allows to use basic tensor operations in standard ML frameworks and benefit from the GPU.
This allows to handle millions of elements in seconds.
The method and the implementation are also agnostic to the input dimension $D$.
However, the use in $D>8$ is impractical even for small networks due to the exponential growth of the complex (see Figure \ref{fig:counts}).

Our contributions are summarized as follows:
\begin{itemize}
    \item A novel method to extract the polyhedral complex of a \gls{ReLU} \glspl{NN} in general dimensions with a focus on performance.
    \item A novel set of experiments directly optimizing the geometric properties of the complex enabled by the fast and differentiable access to the polyhedral complex.
    \item An open source implementation\footnote{\href{https://github.com/arturs-berzins/relu_edge_subdivision}{github.com/arturs-berzins/relu\_edge\_subdivision}} using standard tensor operations in \texttt{PyTorch} and leveraging the GPU.
\end{itemize}

\def\width{\textwidth}
\begin{figure*}
\centering
\includegraphics[width=\width]{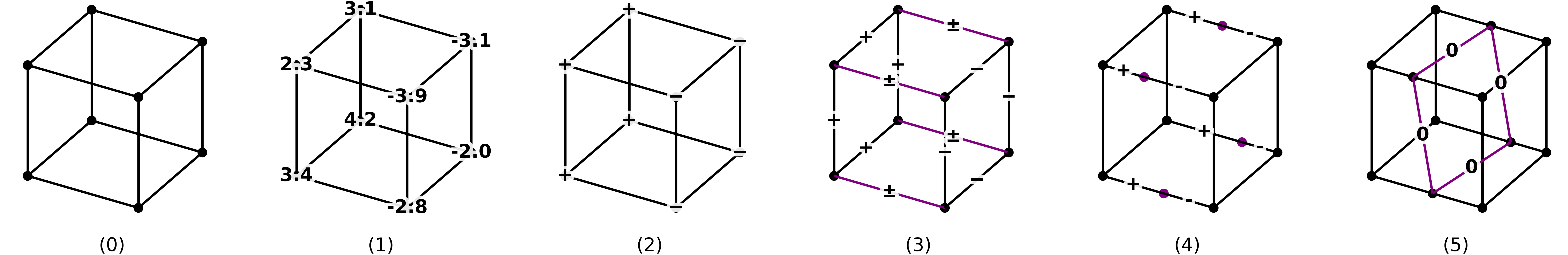}
\caption{
Steps of a single iteration of \emph{edge subdivision}.
Starting with the current 1-skeleton (0), evaluate the \gls{NN} at the vertices (1) and determine the sign of the relevant neuron (2). %
If the signs of a vertex pair sharing an edge differ, the hyperplane must intersect this edge (3).
This intersection is a new vertex whose location interpolates the coordinates and values of the vertex pair and splits the edge in two (4).
To build new edges, connect the new vertices sharing a face (5).
}
\label{fig:steps}
\end{figure*}

\section{Related work}
\label{sec:related}

\paragraph{\gls{CPWA} \glspl{NN}}
A \gls{NN} is itself a \gls{CPWA} function if it is a composition of affine operators, such as fully-connected layers, convolutional layers, skip connections and \gls{CPWA} activation functions, such as (leaky) ReLU, absolute value, hard hyperbolic tangent, hard sigmoid, and max-pooling. %
Many previous authors have investigated \gls{CPWA} \glspl{NN} and fully-connected \glspl{NN} with \gls{ReLU} activation in particular.
Examples include the study of their expressivity in terms of the number of affine regions, \cite{Pascanu2013, Montufar2014, Telgarsky2015, Telgarsky2016, Raghu17, Serra2018, Hanin2019, Sattelberg2020, Wang22estimation}, their connection to adversarial robustness \cite{Jordan2019, Hein2019, Balint22gaussian}, max-affine spline operators, vector quantization, and K-means clustering \cite{Balestriero2021}, batch-norm \cite{Balestriero2022BatchNE}, affine constraint enforcement \cite{Balastriero2022Police}, reverse-engineering \cite{rolnick20reverseengineering} and the geometry of the regions \cite{Balestriero2019, Balestriero2021, Grigsby22BentHyperplane}.

\paragraph{Number of regions}
The maximum number of regions is known to be polynomial in the width and exponential in the depth and input dimension of the \gls{NN} \cite{Raghu17, Montufar2014}.
In practice, however, both randomly initialized and trained \glspl{NN} have a number of regions which is much smaller, with the growth being only polynomial in the number of neurons, but still exponential in the number of input dimensions \cite{Hein2019, Hanin2019DeepReLU, Hanin2019} making the problem of counting regions NP-hard \citet{Wang22estimation}.
As a consequence, for high-dimensional input-spaces even very small \glspl{NN} have an extremely large number of regions, making their enumeration difficult.

\citet{Serra2018} devise a mixed-integer linear program to count the number of regions in a \gls{ReLU}-network with an arbitrary input dimension.
It is demonstrated that a \gls{NN} trained on MNIST with 784-dimensional input and a total of just 22 hidden neurons generates $O(10^7)$ regions which takes tens of hours to count on a server-grade machine.

\subsection{Complex extraction}
Many works provide illustrations of the regions of a 2D input space, which can be acquired by determining the neuron states at points sampled on the image grid.
While this serves as an approximation, there are two known methods operating on the exact complex: region subdivision and marching. 

\paragraph{Region subdivision}
Several works describe \cite{Raghu17} and implement \cite{Hanin2019, Wang22estimation, humayun2022exact, humayun2023splinecam} region subdivision as a method to extract the exact polyhedral complex from a \gls{ReLU}-network.
Starting with an initial polytope, the idea is to sequentially consider each neuron of each layer.
For each neuron calculate the affine map on every existing polytope and determine whether the hyperplane cuts the region in two.
Our method builds upon this interpretation, but, instead of the regions, subdivides the edges to solve the redundancy in neighbouring regions.

\paragraph{Marching}
\citet{Lei2021} propose \emph{Analytical Marching} to extract the 0-isosurface of a \gls{CPWA} neural implicit shape with a bounded \gls{CPWA} 2D boundary in 3D space.
The algorithm is initialized by identifying a point on the 0-isosurface and the corresponding activation pattern or state of the face.
Each edge of a face is the intersection of the face plane and a boundary plane induced by the affine map of some other neuron in the \gls{NN}.
Consequentially, a vertex of a face is the intersection of the face plane and two boundary planes.
However, not all potential edges and vertices are valid, so it is checked whether they have the same state, i.e., whether they lie on the same side of all boundary planes as the face itself.
Each valid edge is then used to pivot to a neighboring face by flipping the activation corresponding to the edge neuron.
Analytical Marching serves as an exact alternative to classic mesh extraction methods, such as marching cubes, offering a trade-off between precision and performance.
However, it is unclear how the method generalizes to the full volumetric complex and higher dimensions.

\section{Background}
\label{sec:Background}

It is well known that each of the regions supporting the \gls{CPWA} \gls{NN} is an intersection of affine halfspaces forming a convex polyhedral set.
Together they partition the input space into a polyhedral complex \cite{Balestriero2019, Hanin2019, Grigsby22BentHyperplane}.

We start by introducing the relevant terminology from the classic theory on polyhedral complexes and hyperplane arrangements.
We then generalize to folded hyperplane arrangements due to \gls{CPWA} \glspl{NN}.
Lastly, we discuss the intersection-poset and sign-vectors as a means to exploit the combinatorial structure of the folded hyperplane arrangement.

\subsection{Polyhedral complexes}
We start by reviewing select facts about polyhedral complexes and refer to a more thorough treatment of the topic in the context of geometry \cite{Grunbaum2003ConvexPolytopes, Grunert2012morse} and ReLU networks \cite{Grigsby22BentHyperplane}.
\\
A \emph{hyperplane} $H := \left\{ \x \in \R^D | \w^\top\x - b = 0 \right\}$ is the zero-level set of an affine map with the slopes $\w\in\R^D$ and a threshold $b\in\R$. 
We will assume that all hyperplanes are \emph{non-degenerate}, meaning $\w \neq \mathbf{0}$.
The sublevel set $H^-$ and the super-level set $H^+$ are the negative and positive \emph{half-spaces}, respectively.
\\
A \emph{polyhedral set} $\mathcal{P}$ in $\R^D$ is the closure of an intersection of finitely many half-spaces $H^+_1,...,H^+_m \subseteq \R^D$.
This is called the \emph{H-representation} of $\mathcal{P}$. %
The \emph{dimension} of the polyhedral set $\mathcal{P}$ is the dimension of its affine hull.
\\
A hyperplane $H$ in $\R^D$ is a \emph{cutting hyperplane} of $\mathcal{P}$ if there exist $\x_1, \x_2 \in \mathcal{P}$ with $\x_1\in\mathcal{P}\cap H^+$ and $\x_2\in\mathcal{P}\cap H^-$.
A hyperplane $H$ in $\R^n$ is a \emph{supporting hyperplane} of $\mathcal{P}$ if $H \cap \mathcal{P} \neq \emptyset$ and $H$ does not cut $\mathcal{P}$.
\\
The intersection $F = H \cap \mathcal{P}$ is a \emph{face} of $\mathcal{P}$ for some supporting hyperplane $H$ of $\mathcal{P}$.
$F=\emptyset$ and $F=\mathcal{P}$ are \emph{improper} faces of $\mathcal{P}$, otherwise $F$ is \emph{proper}.
A \emph{$k$-face} of $\mathcal{P}$ is a face of $\mathcal{P}$ of dimension $k$. A 0-face is a \emph{vertex}, 1-face is an \emph{edge}, and $(D-1)$-face is a \emph{facet}. %
If $\mathcal{P}$ is bounded, its \emph{V-representation} is its set of vertices with the convex hull $\mathcal{P}$.
\\
A \emph{polyhedral complex} $\mathcal{C}$ of dimension $D$ is a finite set of polyhedral sets of dimension $k=0..D$, called the \emph{cells} of $\mathcal{C}$, such that 
(i) if $C\in\mathcal{C}$ then every face of $C$ is in $\mathcal{C}$;
(ii) if $B,C\in\mathcal{C}$ then $B \cap C$ is a single mutual face of both $B$ and $C$.
\\
The \emph{domain} of $\mathcal{C}$, denoted $|\mathcal{C}|$, is the union of its cells.
Conversely, we call $\mathcal{C}$ the \emph{polyhedral decomposition} of the domain $|\mathcal{C}|$.
\\
A \emph{polyhedral subcomplex} of $\mathcal{C}$ is a subset $\mathcal{C}' \subset \mathcal{C}$ such that for every cell $C$ in $\mathcal{C}'$, every face of $C$ is also in $\mathcal{C}'$.
The \emph{$k$-skeleton} of $\mathcal{C}$, denoted $\mathcal{C}_k$, is the subcomplex of all cells of $\mathcal{C}$ of dimension $i=0..k$.

\subsection{Hyperplane arrangements}
A \emph{hyperplane arrangement} is a finite set of hyperplanes $\mathcal{H} = \left\{ H_1,\dots,H_m \right\}$ in $\R^D$.
It induces a polyhedral decomposition $\mathcal{C}(\mathcal{H})$ of $\R^D$.
A $D$-dimensional cell in $\mathcal{C}(\mathcal{H})$ or a \emph{region} is the closure of a maximal connected region of $\R^D$ not intersected by any hyperplane in $\mathcal{H}$.
For $k=0..D-1$ the $k$-dimensional cells in $\mathcal{C}(\mathcal{H})$ are defined inductively as the facets of the $(k+1)$-dimensional cells.
A hyperplane arrangement is \emph{generic} if no more than $D$ hyperplanes intersect at any single point.

\subsection{Sign-vectors}
Given a hyperplane arrangement $\mathcal{H}$, any point $\x \in \R^D$ is assigned a \emph{sign-vector} $\bm{\sigma}(\x) = (\sigma_i(\x))_{i=1..m}$, with 
\begin{equation}
    \sigma_i(\x) = \begin{cases}
        + &\text{if } \x \in H_i^+,
        \\
        0 &\text{if } \x \in H_i,
        \\
        - &\text{if } \x \in H_i^-.
    \end{cases}
\end{equation}
Similarly, every cell $C$ of $\mathcal{C}(\mathcal{H})$ can be associated with a sign-vector $\bm{\sigma}(C)$ such that
\begin{equation}
C = \bigcap_{i=1}^m H_i^{\sigma_i(C)} =: \mathcal{H}^{\bm{\sigma}(C)}    
\end{equation}
with $H^0:=H$ \cite{Matousek2002discrete}.

\subsection{ReLU networks and folded hyperplane arrangements}

For our purposes a fully-connected feed-forward \gls{NN} $f_{\Theta}$ maps any point $\x$ in the \emph{domain} $\mathcal{D} \subset \R^{D}$ to a $D^{(L)}$-dimensional output $f_\Theta (\x) \in \R^{D^{(L)}}$.
The \gls{NN} is a composition of $L$ layers with parameters $\Theta = \left\{ \Theta^{(l)} \right\}_{l=1..L}$: %

\begin{equation}
    f_{\Theta}(\x) = \paren{ 
        f^{(L)}_{\Theta^{(L)}} \circ
        \dots
        \circ f^{(1)}_{\Theta^{(1)}}
    }(\x) \ .
\end{equation}

Starting at $\x^{(0)}=\x$, the layers are applied successively for $l=1..L$ as
\begin{equation}
    \x^{(l)} = f^{(l)}_{\Theta^{(l)}} \paren{\x^{(l-1)}} = \ReLU \paren{ \W^{(l)} \x^{(l-1)} + \b^{(l)} } .
\end{equation}

The layer parameters $\Theta^{(l)} = \left\{ \W^{(l)}, \b^{(l)} \right\}$ contain the \emph{weights} $\W^{(l)} \in \mathbb{R}^{D^{(l-1)} \times D^{(l)}}$ and \emph{biases} $\b^{(l)} \in \mathbb{R}^{D^{(l)}}$.
For simplicity, we adhere to the main line of work focusing on the use of $\ReLU(x) = \max(0, x)$ as the activation, but the key ideas generalize to any \gls{CPWA} \gls{NN}.

In analogy to a hyperplane which is the zero-level set of an affine map, a \emph{folded hyperplane} is the zero-level set of the pre-activation of a neuron.
The $i$-th neuron in the $l$-th layer induces the folded hyperplane $H_i^{(l)} := \left\{ \x \in \R^D | \W^{(l)\top}_i \x^{(l-1)} + b^{(l)}_i = 0 \right\}$.
Similarly, a finite set of folded hyperplanes $\mathcal{H}$ is a \emph{folded hyperplane arrangement} and induces a polyhedral decomposition of the domain $\R^D$ \cite{Hein2019, Grigsby22BentHyperplane}.
On each region, the folded hyperplane acts like an affine hyperplane and does not fold.
The sign-vector is defined analogously and can be evaluated from the neuron pre-activation: $ \sigma^{(l)}_i (\x) = \operatorname{sgn}( \W^{(l)\top}_i \x^{(l-1)} + b^{(l)}_i ) $.

Theorem 3 in \citet{Grigsby22BentHyperplane} states that almost every \gls{ReLU} network is generic, which we assume throughout this work.

\subsection{Polyhedral combinatorics}
\label{sec:combinatorics}

A partially ordered set or \emph{poset} is the pair $(\mathcal{S}, \leq)$ of the set $\mathcal{S}$ together with a binary relation $\leq$ on $\mathcal{S}$ (called an \emph{ordering}) satisfying three axioms: reflexivity ($x \leq x$ for all $x$), transitivity ($x \leq y$ and $y \leq z$ implies $x \leq z$), and weak anti-symmetry (if $x \leq y$ and $y \leq x$, then $x = y$).
For any two elements $x, y \in \mathcal{S}$ the \emph{meet} $x \land y$ is the greatest lower bound of $x$ and $y$.
Similarly, the \emph{join} $x \lor y$ is the least upper bound of $x$ and $y$.
Neither need exist, but if they do then they are unique \cite{Kishimoto2019PolyhedralPO}.

We will introduce some terminology from graph theory to denote relationships in the poset.
$y\in\mathcal{S}$ is an \emph{ascendant} of $x\in\mathcal{S}$ if $x<y$.
Conversely, $x$ is a \emph{descendant} of $y$.
The closest common ascendant of both $x$, $y$ is the join $x \lor y$.
The closest common descendant of both $x$, $y$ is the meet $x \land y$.
$y\in\mathcal{S}$ is a \emph{parent} of $x\in\mathcal{S}$ if $x<y$ and no $z\in\mathcal{S}$ satisfies $x<z<y$.
Conversely, $x$ is a \emph{child} of $y$.
\\
The \emph{intersection-poset} of the polyhedral complex $\mathcal{C}$ is the poset $(\mathcal{C}, \subseteq)$ of its cells ordered by inclusion.

\section{Method}
\label{sec:method}

Our method is motivated by the observation illustrated in Figure \ref{fig:continuous}.
Due to the continuity of the activation function, all folded hyperplanes are continuous across each other.
However, the existing subdivision methods consider each region independently.
As a consequence, upon the intersection with a new folded hyperplane, each new vertex is computed independently $2^{D-1}$ times in V-representation, since an edge has $2^{D-1}$ ascendant regions in an unbounded arrangement.
Similarly, in H-representation, the hyperplane redundancy check performed via linear programming arrives at the same conclusion on all $2^{D-1}$ ascendant regions of the shared edge.
\\
Our method alleviates this redundancy by taking into account the continuity and disregarding the regions, instead using only the unique vertices and edges, i.e. the 1-skeleton.
Edge subdivision preserves the iterative structure of considering each neuron in each layer sequentially, for each neuron subdividing the edges in five steps:
\begin{itemize}
    \itemsep=0em
    \item [(1)] Evaluate the \gls{NN} at the vertices;
    \item [(2)] Get the sign-vectors of the vertices;
    \item [(3)] Find splitting edges by comparing the signs of vertex pairs;
    \item [(4)] For each splitting edge, compute the new vertex using interpolation and split the edge;
    \item [(5)] Build the intersecting edges (connecting new vertices across splitting faces).
\end{itemize}
This process is illustrated in Figure \ref{fig:steps} and the steps are detailed in the following.
We start by discussing how to recover the combinatorial structure of the complex from the sign-vectors.

\def\width{0.8\columnwidth}
\begin{figure}
\begin{center}
\includegraphics[width=\width]{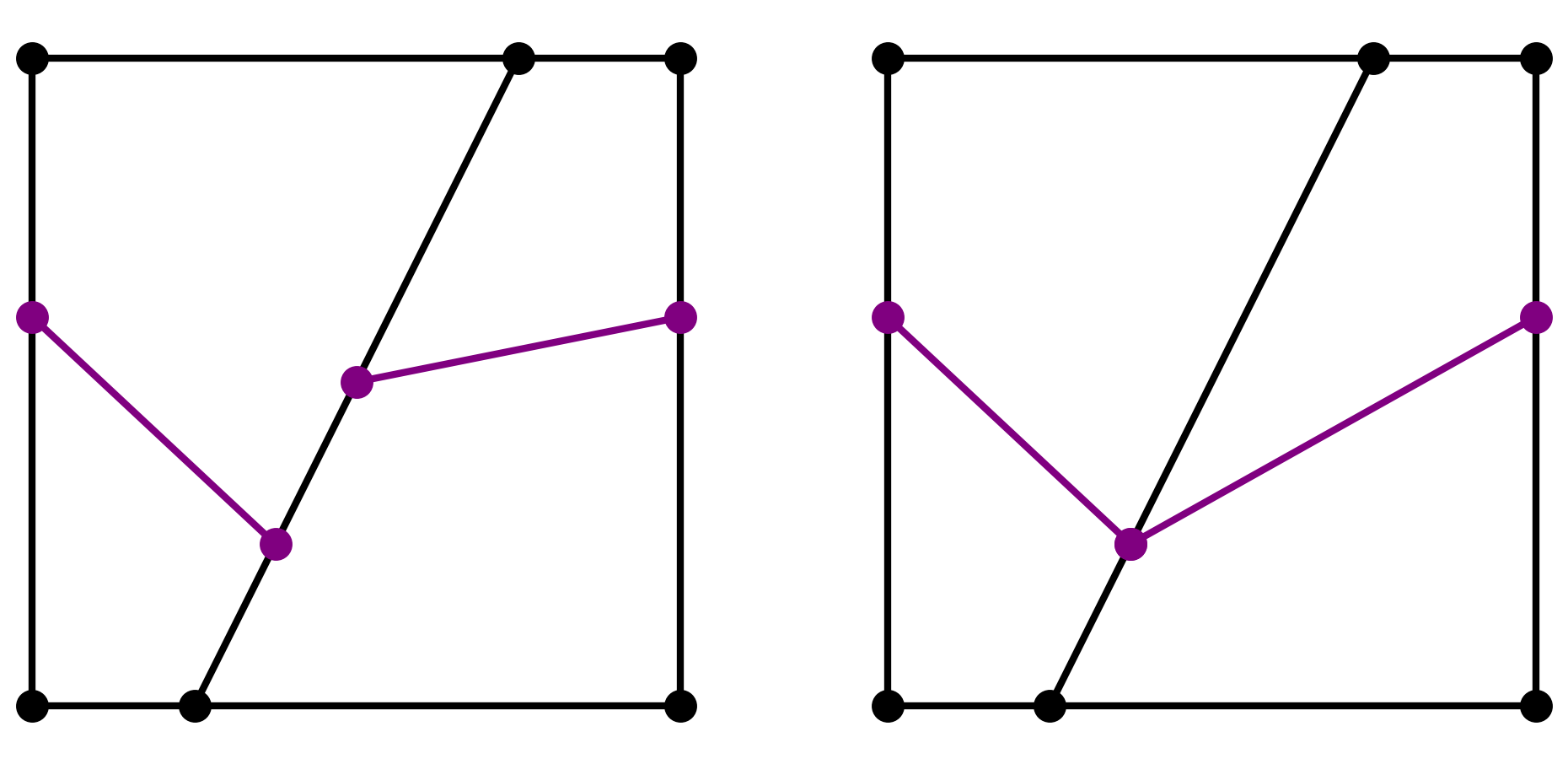}
\end{center}
\caption{Motivation in $D=2$: due the continuity of the activation, the two new edges share the same vertex on the common edge of the two regions. Processing each region individually is redundant.
}
\label{fig:continuous}
\end{figure}

\def\width{0.5\columnwidth}
\begin{figure}
\begin{center}
\includegraphics[width=\width]{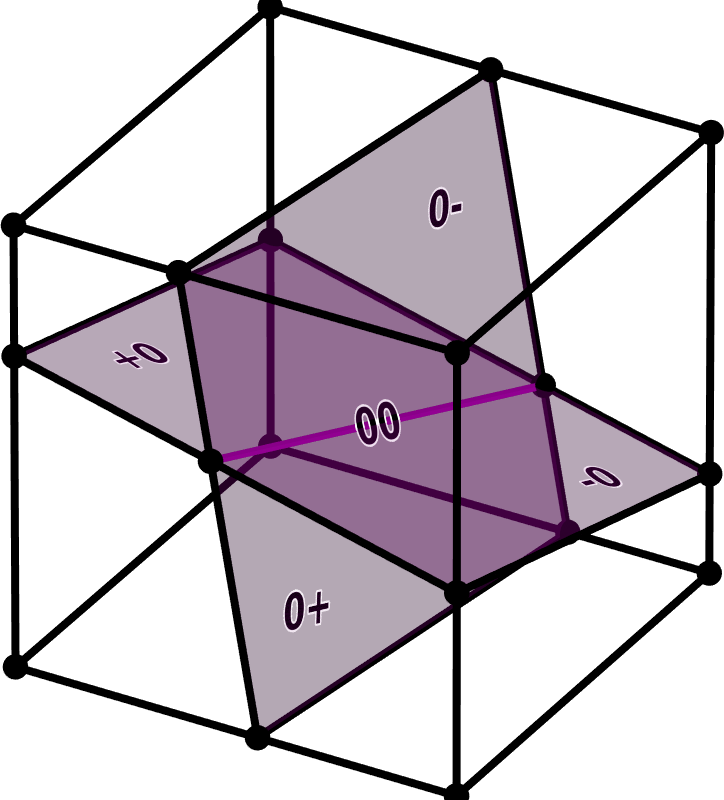}
\end{center}
\caption{
The parenting $(k+1)$-faces of a $k$-face can be obtained by perturbing each zero in its sign-vector at a time. 
Here, $k=1$ and the first $m=6$ entries of the sign vector are hidden for visual clarity since they are all $+$.
This edge and all its ascendants are interior cells.
}
\label{fig:perturbation}
\end{figure}

\subsection{Perturbation using sign-vectors}
\label{sec:operations}

All the combinatorial relationships described in Section \ref{sec:combinatorics} can be easily evaluated using sign-vectors.
However, instead of just determining the relationship of two given cells, edge subdivision and optional post-processing steps rely on building parent cells.

For now, assume the unbounded domain $\R^D$.
The number of zeros in the sign-vector of a $k$-cell is $(D-k)$.
To construct all the parents of this cell, take one of the zeros at a time and set it to $+$ or $-$.
Consequentially, a $k$-cell has $2(D-k)$ parent cells for $k=0..D-1$.
We call this process \emph{perturbation} and illustrate it in Figure \ref{fig:perturbation}.
\\
Repeating this for all $k$-cells in the complex constructs all $(k+1)$-cells, including the ordering relations.
Hence, the $(k+1)$-skeleton can be built from the $k$-skeleton.
Starting from the 1-skeleton, perturbation can be applied sequentially to reconstruct the whole intersection-poset, including the regions.

Instead of the unbounded $\R^D$, we will operate on a bounded polyhedral domain $\mathcal{D}$, which is given by the intersection of $m$ affine halfspaces.
The motivation for this is a simpler implementation since bounded edges have exactly two vertices allowing to use simpler data-structures.
In this setting, special care must be taken with \emph{boundary cells}.
These are the cells for which any of the first $m$ entries of the sign-vector is $0$ (conversely, the first $m$ entries of an interior cell are $+$).
Since we are not interested in cells outside the domain, we perturb the zeros on the boundary only toward the interior $+$.
So under the consideration of the bounded domain, a $k$-cell has $z + 2(D-k-z)$ parent cells, where $z$ is the number of zeros in the first $m$ signs.

\subsection{Edge subdivision}

To understand how the 1-skeleton is subdivided, i.e. how new $0$- and $1$-cells are created, consider a new hyperplane $H$ cutting a $k$-cell $C$. 
Their intersection $C^0 = C \cap H$ is a new $(k-1)$-cell.
Additionally, $H$ splits $C$ into two new $k$-cells $C^+ = C \cap H^+$ and $C^- = C \cap H^-$.
So there are exactly two mechanisms for creating $(k-1)$-cells: (i) splitting a $(k-1)$-cell with $H$ which preserves the dimension and (ii) intersecting a $k$-cell with $H$ which lowers the dimension.
Focusing on $k=0,1,2$ as sources for new $0,1$-cells leads us to edge subdivision.

\subsubsection{Steps 0-4}

Let $\mathcal{V}$ and $\mathcal{E}$ be the set of all vertices and edges at the current iteration (step 0).
Let $H$ be the next folded hyperplane to intersect with and recall that it behaves like an affine hyperplane on each region, folding at its facets.
Per generality assumption, no vertex in $\mathcal{V}$ will intersect $H$.
Each vertex in $\mathcal{V}$ is $+$ or a $-$ sign w.r.t. $H$.
These signs can be determined from the pre-activation values of the neuron corresponding to $H$, obtained by simply evaluating the \gls{NN} at the vertices (steps 1, 2).

We call $E\in\mathcal{E}$ a \emph{splitting} edge if $H$ cuts $E$.
Splitting edges can be identified by their two vertices having opposite signs, which we label $V^+, V^-$ (step 3).
The new vertex $V^0=E \cap H$ on the splitting edge $E$ can be computed by linearly interpolating the positions of $V^+, V^-$ weighted by their pre-activation values. %
This new vertex takes the sign $0$ w.r.t. $H$.
The old splitting edge $E$ is removed from $\mathcal{E}$ and the two new \emph{split} edges $E^+ = E \cap H^+$ with vertices $V^+, V^0$ and $E^- = E \cap H^-$ with vertices $V^-, V^0$ are added to $\mathcal{E}$.
The new signs of $E^+, V^+$ and $E^-, V^-$ w.r.t. $H$ are trivially $+$ and $-$, respectively (step 4).

\subsubsection{Step 5}

This completes intersecting and splitting edges with the folded hyperplane.
However, new edges are also formed where $H$ intersects 2-faces.
We call $F$ a \emph{splitting} 2-face if $H$ cuts $F$.
We call their intersection $E^0 = F \cap H$ an \emph{intersecting} edge.

In a naive approach, it would seem that we need to track the 2-faces in order to intersect them with $H$.
However, by induction, this would require to track the whole complex.
This is impractical due to the amount and layout of the memory -- for $k>1$, a $k$-cell can have an arbitrary number of facets, as opposed to exactly 2 vertices for each edge in a bounded domain.
Instead, we propose to intersect the 2-faces \emph{implicitly}. %

This is enabled by another observation -- every bounded splitting 2-face has exactly two splitting edges. 
Furthermore, the intersecting edge connects the two new vertices on those two splitting edges.
Since we have already determined the splitting edges, we use them to implicitly identify splitting 2-faces and append the intersecting edges to $\mathcal{E}$.

Given two splitting edges we can perform a simple adjacency check using their sign vectors.
However, checking all possible pairs has a quadratic memory complexity $\mathcal{O}(|\mathcal{\hat{E}}|^2)$ in the number of splitting edges $|\mathcal{\hat{E}}|$.
This is infeasible even for moderately sized \gls{NN} (see Section \ref{sec:counting}).

Instead, we propose a much more efficient method.
For each splitting edge build its parenting 2-faces using perturbation as described in Section \ref{sec:operations}.
We have a list of splitting 2-faces each pointing to a single splitting edge.
In this list, each 2-face comes up exactly twice.
We pair the two edges associated with the same 2-face.
With at most $2(D-1)$ 2-faces per edge, the memory requirement is only $\mathcal{O}(2(D-1)|\mathcal{\hat{E}}|)$.
Lastly, it remains to add the intersecting edge to $\mathcal{E}$. 
Its sign-vector is inherited from the splitting face with a $0$ appended w.r.t. $H$. 

This concludes a single iteration of edge subdivision, which is repeated for every neuron in every layer.

\subsubsection{Complexity analysis}
\label{sec:complexity}

In Appendix \ref{app:complexity}, we elaborate that all the steps (1-5) of edge subdivision can be implemented in linear time and memory complexity in the number of vertices $|\mathcal{V}|$, edges $|\mathcal{E}|$, or splitting edges $|\hat{\mathcal{E}}|$.
We argue further that $O(|\mathcal{E}|)$ and $O(|\hat{\mathcal{E}}|)$ can be replaced with $O(|\mathcal{V}|)$, concluding that the total algorithm is $O(|\mathcal{V}|)$, and hence optimal.

The number of regions in a randomly initialized or trained \gls{NN} is known to be $o(N^D/D!)$ where $N$ is the total number of neurons \cite{Hanin2019DeepReLU}.
Conservatively assuming a proportional number of vertices $|\mathcal{V}|$ (see Figure \ref{fig:counts}), we can obtain the complexity of the algorithm with respect to the \gls{NN} architecture.

\section{Experiments}
\label{sec:experiments}

We start by describing, validating, and timing our implementation of edge subdivision.
As described in Sections \ref{sec:introduction} and \ref{sec:related}, the access to the exact polyhedral complex is intriguing in many theoretical and practical applications.
Instead, we consider how the speed and differentiability of our method enable a novel experiment in which an optimization objective is formulated on the geometric properties of the extracted complex.
Lastly, we discuss an approach to pruning \gls{NN} parameters and test a method to modify edge subdivision if the goal is to extract just an iso-level-set, i.e. decision boundary. 

\begin{figure}
\begin{center}
  \includegraphics[width=\columnwidth]{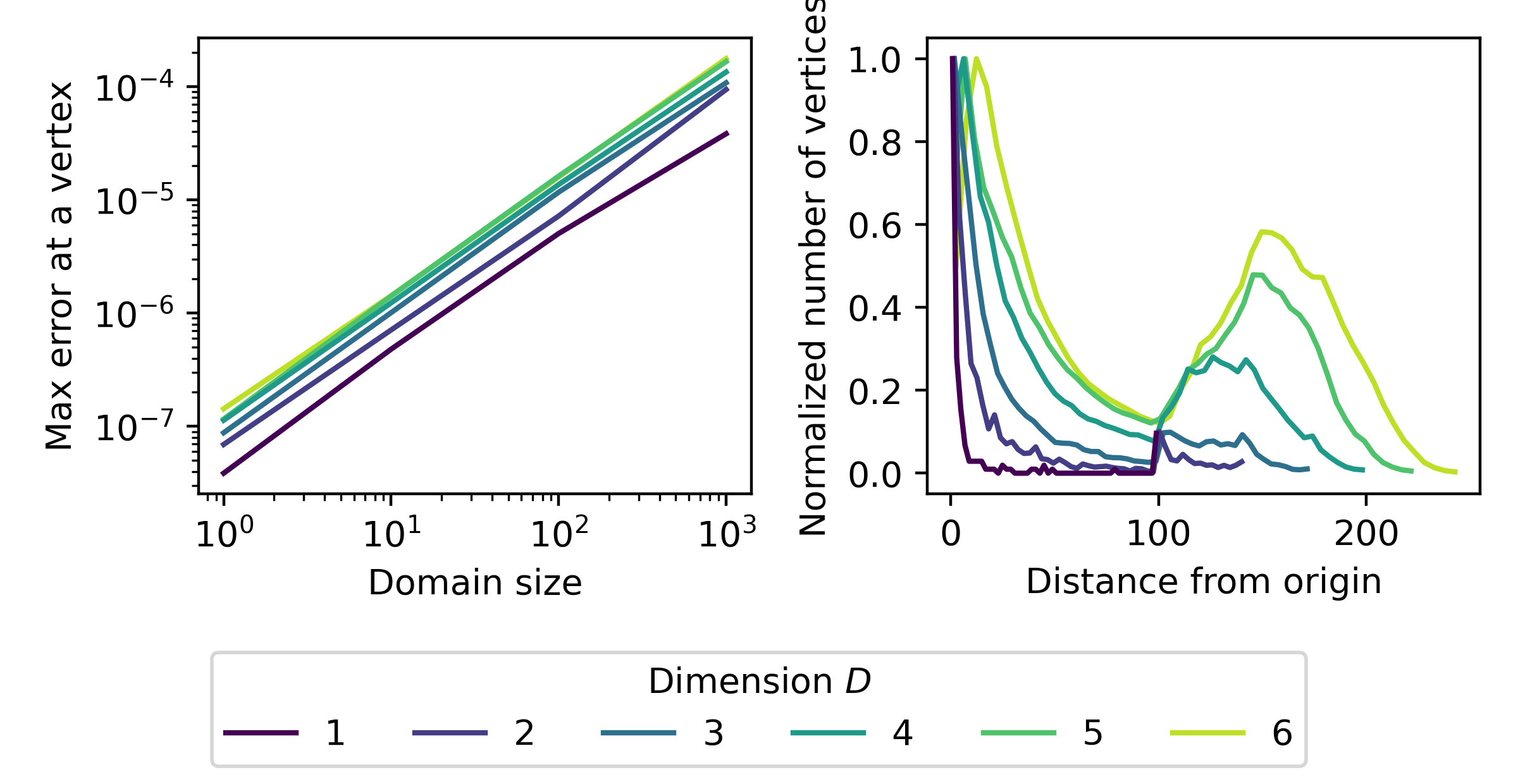}
\end{center}
\caption{
Left: Maximum error over all vertices is at least seven orders of magnitude smaller than the size of the hypercube domain serving as a validation.
Right: Distribution of the vertex distances from the origin, normalized by the maximum value.
The two distinct modes correspond to the interior and boundary vertices.
}
\label{fig:combined}
\end{figure}

\subsection{Implementation}
\label{sec:implementation}
We implement the algorithm in \texttt{PyTorch}.
Since only vertex positions and bounded edges with exactly two vertices are stored, edge subdivision can run efficiently and exclusively on the GPU.
The steps 0-4 can be implemented using standard tensor operations.
However, using standard operations step 5 can only be implemented in sub-optimal log-linear time using sorting to pair up identical rows of a tensor.
This step can be implemented in linear time using hash-tables, but since efficient hashing on the GPU with custom length keys is non-trivial \cite{Junger2020Hash, Awad2023Hash}, we hope to address this in future work.

\subsubsection{Validation}
\label{sec:validation}
We first test a basic necessary (although insufficient) condition for the validity of our implementation.
For each found vertex, the neurons corresponding to 0 entries in its sign-vector should be 0 at the vertex location.
This is indeed satisfied up to a numerical error.
Figure \ref{fig:combined} (left) shows that the maximum numerical error is roughly seven orders of magnitude smaller than the extent of the domain and even less in lower dimensions.
A fixed \gls{NN} with 4 layer depth and 10 neuron width is used throughout.
This error should not be confused with the geometric error in the position of the vertex, which is closely related, but not quantified here.

\subsubsection{Performance}
\label{sec:counting}

We investigate the time and memory behaviour of our implementation by counting vertices and edges on the bounded unit hypercube domain.
We consider \glspl{NN} of four layers and widths of $10,20,40$ for input dimensions $D=1..10$.
The sizes of the experiments are mainly limited by the memory.
The tests are performed on an NVIDIA RTX 3090. %
Since no data is on the CPU, the maximum allocated memory is measured for just the GPU.%

\begin{figure*}
\begin{center}
  \includegraphics[width=.9\textwidth]{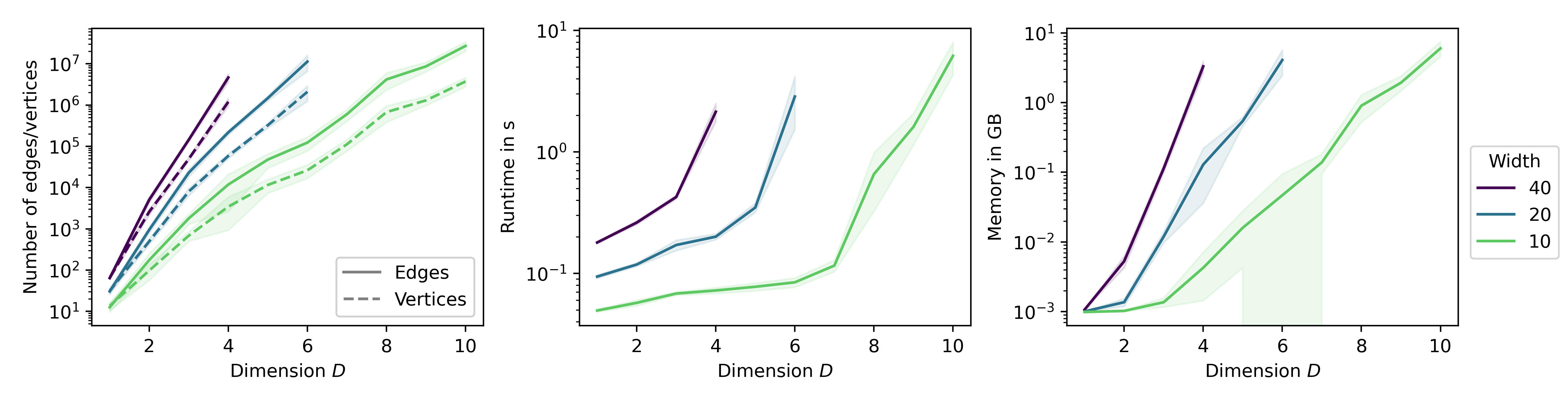}
\end{center}
\caption{
Edge and vertex counts, runtime, and memory usage of randomly initialized \glspl{NN} of four layers and different widths and input dimensions. Mean $\pm$ standard deviation over 5 runs.
}
\label{fig:counts}
\end{figure*}

The results are illustrated in Figure \ref{fig:counts}.
Even moderately sized \glspl{NN} induce complexes with millions of cells, especially as the input dimension increases.
It is known that the number of regions scales exponentially with the number of input dimensions.
For the number of vertices and edges, we observe a subexponential growth.

Counting the regions is possible as described in Section \ref{sec:operations}, but requires an additional significant amount of memory and time, since using perturbation requires to store and group $2(D-1)|\mathcal{E}|$ cells which is up to $\mathcal{O}(10^8)$ for some of the considered cases.

\begin{figure}
\begin{center}
  \includegraphics[width=1.0\columnwidth]{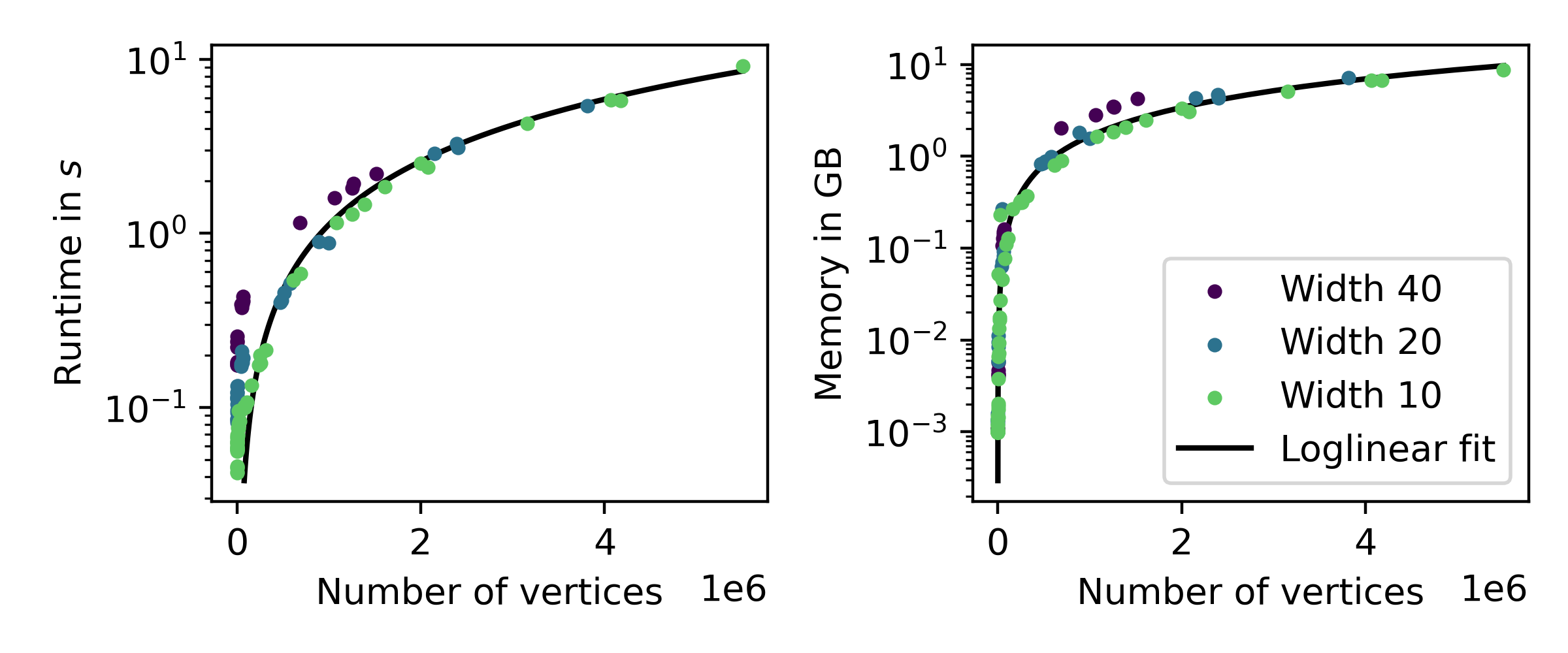}
\end{center}
\caption{
Our implementation shows log-linear scaling w.r.t. the number of vertices.
This is due to a sub-optimal implementation of step 5 using sorting.
Leveraging hash tables would improve the whole implementation to linear complexity.
}
\label{fig:loglinear}
\end{figure}

In Figure \ref{fig:loglinear} we validate our complexity analysis.
We reuse the previous results to plot the runtime and memory over the number of vertices, which as expected is log-linear due to the sub-optimal use of sorting.

\begin{figure}
\begin{center}
  \includegraphics[width=.75\columnwidth]{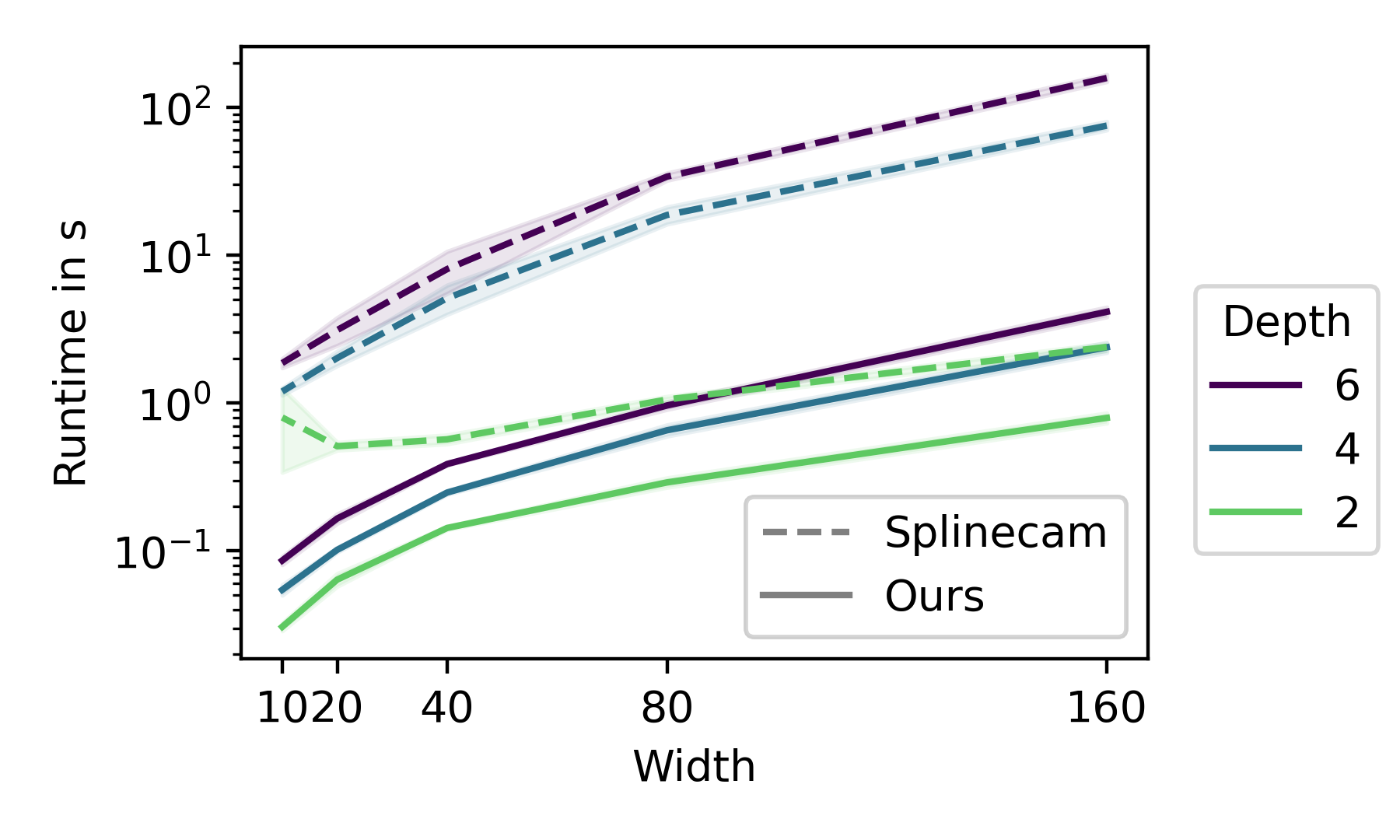}
\end{center}
\caption{
We achieve a 20 times speed-up on average over SplineCam \cite{humayun2023splinecam} which is also limited to $D=2$.}
\label{fig:splinecam}
\end{figure}

Lastly, in Figure \ref{fig:splinecam} we compare to SplineCam \cite{humayun2023splinecam} which is a region subdivision method specifically for $D=2$. 
Over the considered tests, our method is on average 20 times faster, since SplineCam uses graph structures on the CPU.

\subsection{Vertex distribution}
\label{sec:distribution}
We study the effect of the domain being bounded.
We perform edge subdivision on a $\mathcal{D}=[-100,100]^D$ hypercube domain.
For each considered dimension, this is repeated on five randomly initialized \glspl{NN} of 4 layer depth and 10 neuron width.

For every vertex, we compute its distance from the origin $r=||\x||_2$.
Figure \ref{fig:combined} (right) shows a bi-modal distribution of $r$.
For $r<100$, we observe an exponentially decaying density of vertices.
These are the interior vertices due to the folded hyperplane arrangement.
Additionally, there are the domain boundary vertices, which intersect at least one of the hyperplanes defining the domain.
These are located at $r\geq100$, which corresponds to the second mode of the distribution.
For a trained \gls{NN}, we would generally expect a different distribution in the first mode, for example, the vertices to concentrate more tightly around the training data.

This illustrates a limitation of performing edge subdivision on a bounded domain.
If we do not care about the cells on the artificially inserted boundary, then having a large proportion of boundary cells is undesirable for the performance.
One simple solution is to replace the hypercube domain with a simplex domain, whose number of vertices (edges) grows linearly (quadratically) with $D$ as opposed to exponentially.
This also motivates a future work for extending edge subdivision to unbounded domains.

\subsection{Geometric loss}
\label{sec:geometry}

We utilize the differentiability and speed of edge subdivision to consider a novel experiment in which an optimization objective is formulated on geometric properties of the extracted complex.
In Figure \ref{fig:compactness}, we start with a \gls{ReLU} \gls{NN} with two hidden layers of 50 neurons each and $D=2$.
The \gls{NN} is first trained as a neural implicit representation of a bunny.
Then, in each iteration we extract the polyhedral complex and compute the \emph{shape compactness} $c={4 \pi A}/{P^2}$ as the ratio of the area $A$ and the perimeter $P$.
The normalization is such that $c=1$ for the most compact shape -- the circle.
Using $c$ as the loss, the bunny shape converges to a circle in 100 iterations with a standard Adam optimizer.

In general, any geometric loss that depends on the vertex positions can be formulated and optimized, e.g. edge lengths, angles, areas, volumes, curvatures, and other quantities from discrete differential geometry.
This holds for both the boundary and the volumetric shape, as well as the polyhedral complex (i.e. the mesh) itself.

\def\width{0.19\textwidth}
\begin{figure*}
\begin{center}
\begin{tabular*}{\textwidth}{
          @{\extracolsep{\fill}}
        c @{\extracolsep{\fill}}
        c @{\extracolsep{\fill}}
        c @{\extracolsep{\fill}}
        c @{\extracolsep{\fill}}
        c }
  \includegraphics[width=\width]{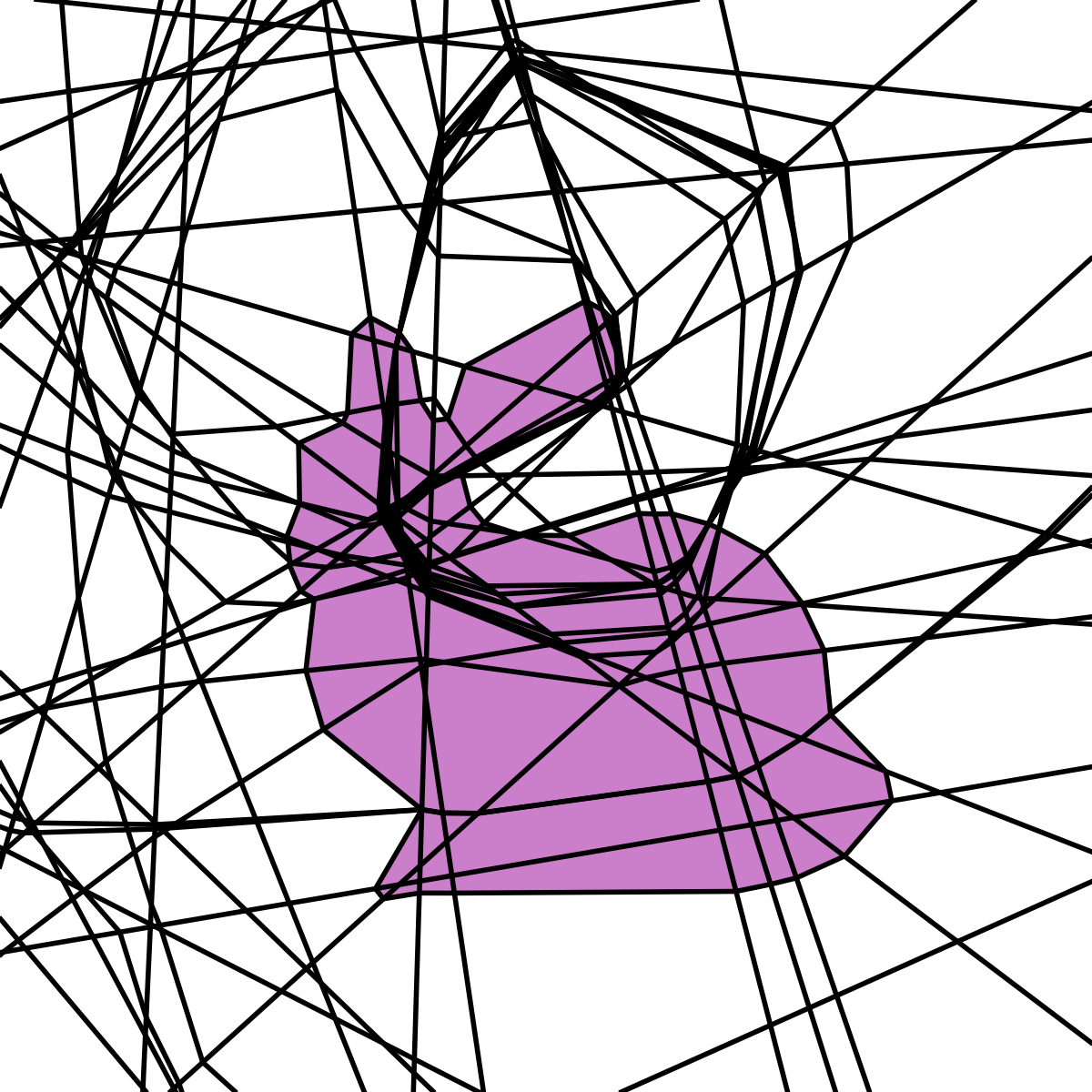} &
  \includegraphics[width=\width]{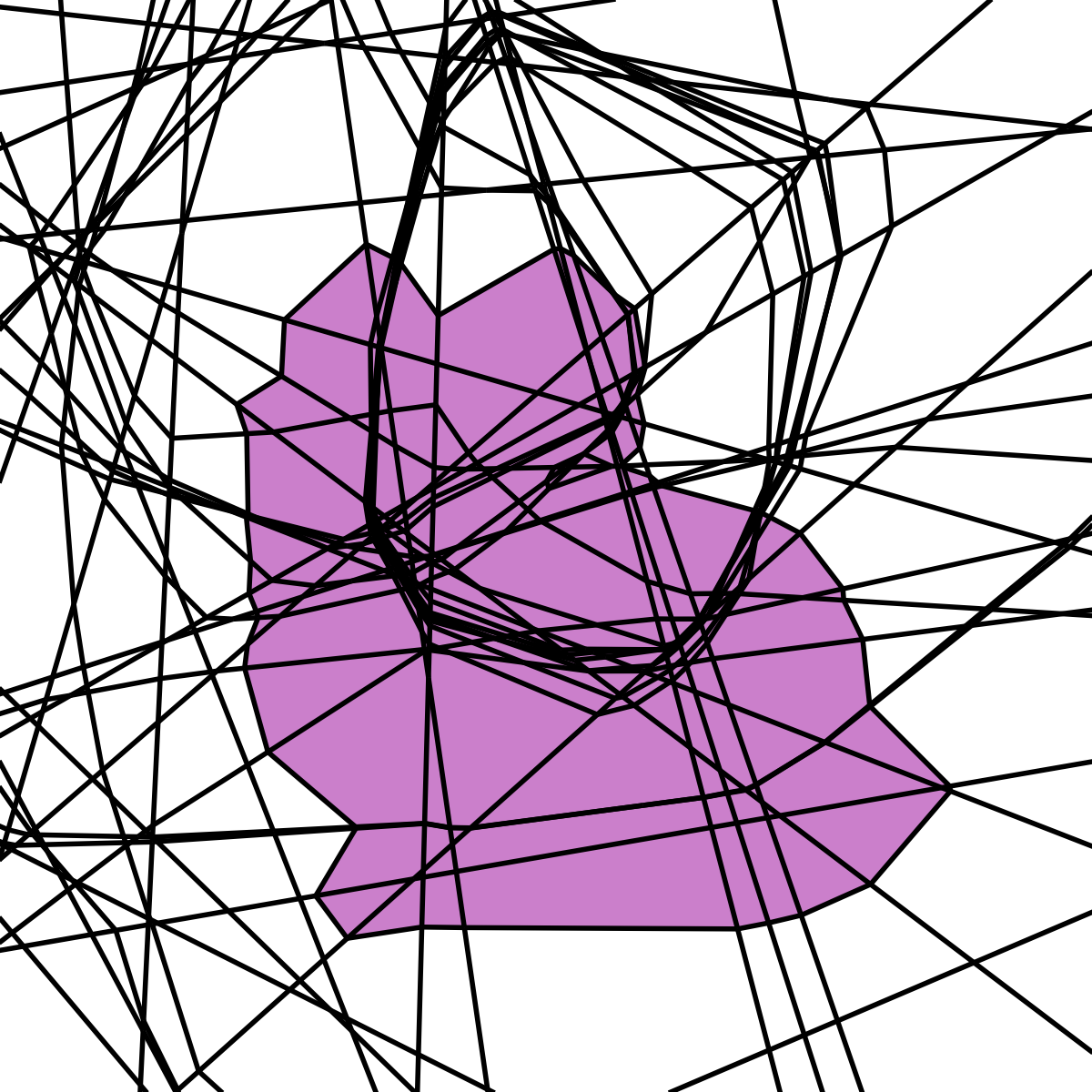} &
  \includegraphics[width=\width]{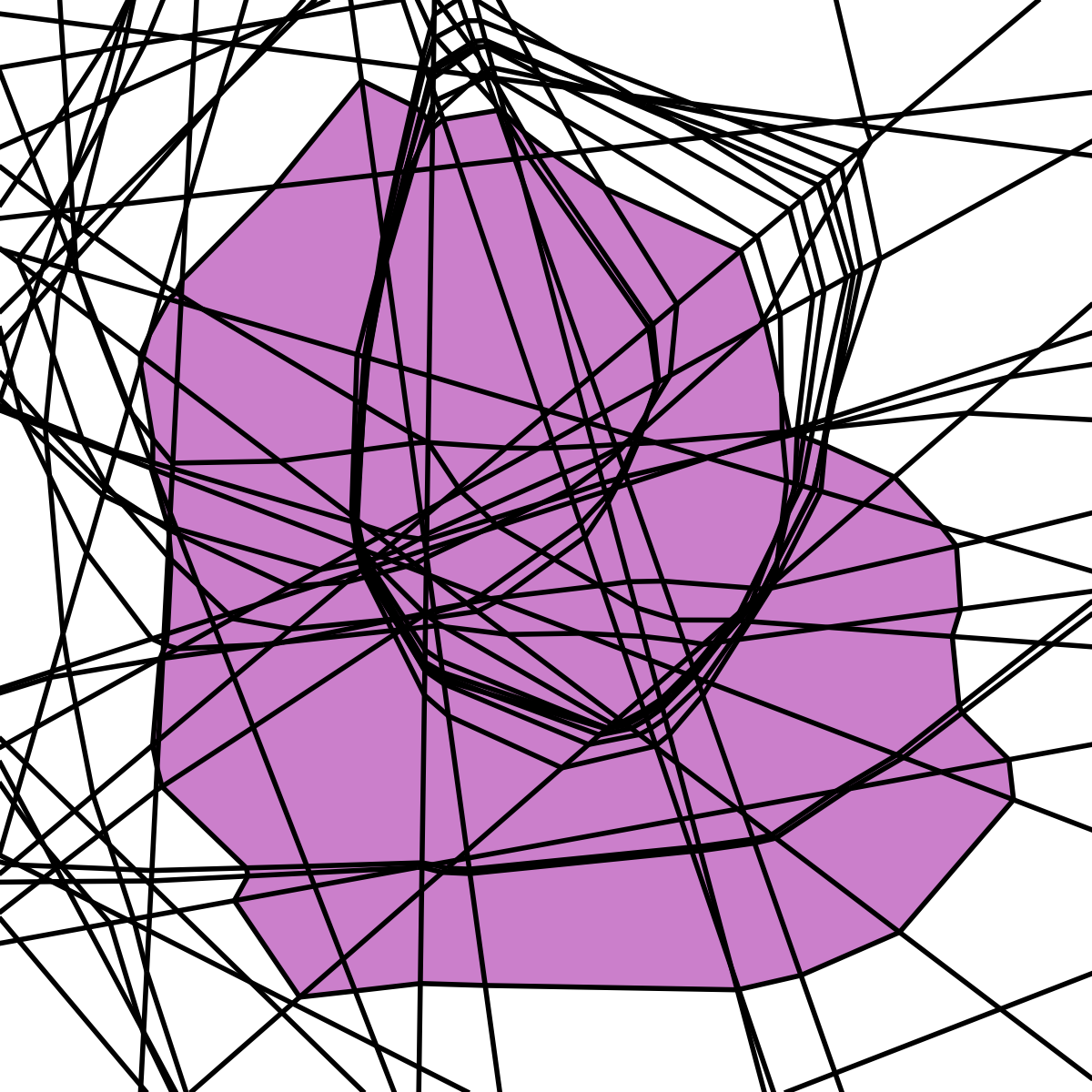} &
  \includegraphics[width=\width]{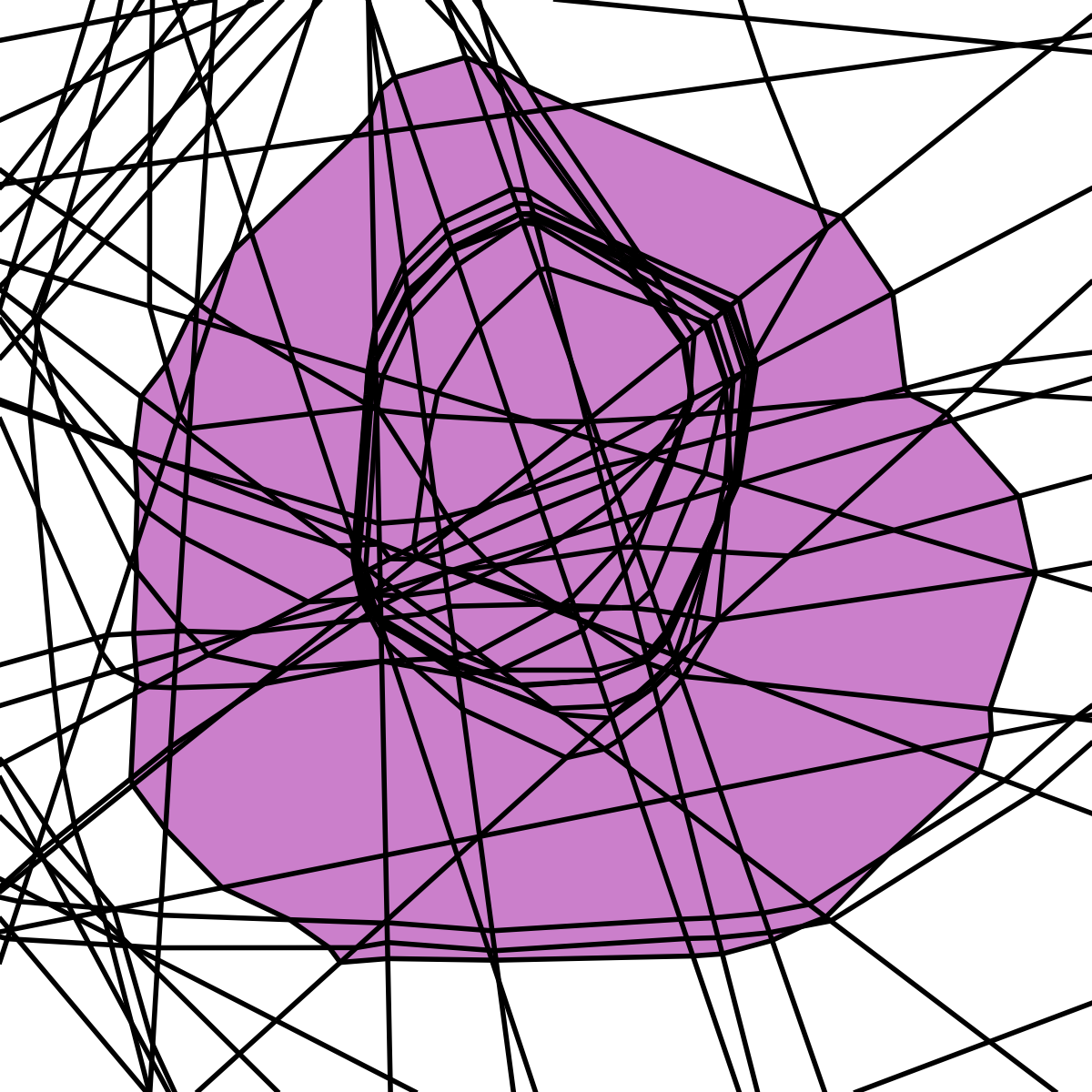} &
  \includegraphics[width=\width]{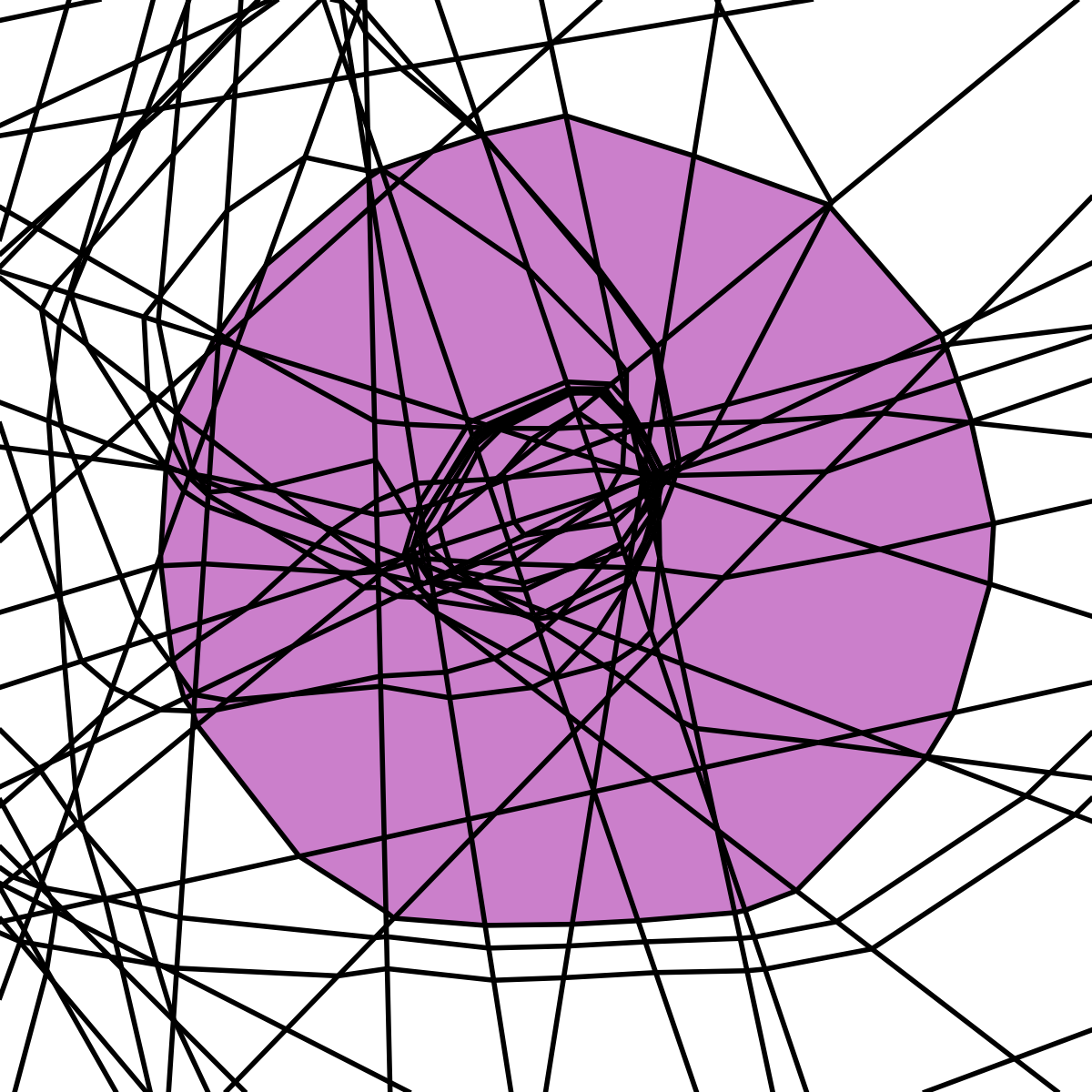}
\end{tabular*}
\end{center}
\caption{
The extracted skeleton at initial, intermediate, and final iteration of optimizing the shape compactness.
The bunny shape converges to a circle in 100 iterations of a standard Adam optimizer with the loss formulated on the extracted complex.
}
\label{fig:compactness}
\end{figure*}

\subsection{Pruning}
\label{sec:pruning}

Finally, we consider two approaches to pruning, focusing on a geometric context due to the intuitive interpretation.
Consider an implicit neural representation of a bounded geometry.
Important in this view is the boundary of the shape, similar to the decision boundary in a classification task.
We can view the \gls{ReLU} \gls{NN} as a compact storage format and many geometric properties of a shape can be computed from just its boundary.

\subsubsection{Parameter pruning}
In the first view, we refer to \emph{pruning} as a \gls{NN} compression technique in which some parameters are removed after training with a negligible drop in the \gls{NN} performance \cite{lee2018snip}.
In the context of preserving the shape or decision boundary, all the folded hyperplanes which do not intersect the boundary can be removed.
This corresponds to pruning the respective neurons. %
For \gls{ReLU}, the boundary can be completely contained in either on the negative or positive half-space of a non-intersecting neuron.
The negative neurons can be removed completely as they do not contribute any value anywhere on the shape.
The folded hyperplanes of such neurons are highlighted in red in Figure \ref{fig:pruning}.
Since the training data is localized on the unit square, the folded hyperplanes not intersecting this domain correspond to \emph{dying-\glspl{ReLU}} -- neurons which for all training samples are in the rectifying 0 region of \gls{ReLU}.
However, there are also folded hyperplanes intersecting the domain but not the shape itself.
Removing all of these allows to compress the $2,50,50,1$ \gls{NN} down to $2,25,19,1$, reducing the number of parameters from $2751$ to $589$.

This can be pruned further by also considering the converse case -- neurons for which the whole boundary is in the linear activation of \gls{ReLU}.
Since each such neuron contributes the same affine function everywhere on the shape, any linearly dependent (in general any $>D$ neurons of the same layer) can be compressed down to $D$ while adjusting the outgoing weights accordingly.

\subsubsection{Pruning during edge subdivision}

Extracting the whole complex and selecting just the boundary is wasteful, even if the \gls{NN} is compressed as described above.
We propose a complementary pruning strategy specifically for during edge subdivision. It intends to prune all edges and vertices, for whom we can say with confidence that they will not contribute toward the boundary.

Recall, that a new vertex is created only where an edge splits and such splitting edges are detected by the signs of their vertex pairs disagreeing.
We can compute the sign-vector of all current vertices even at any intermediate iteration of subdivision.
An edge can be pruned if both its vertices have the same signs w.r.t. all future neurons.
For the considered 2D bunny geometry this reduces the number of edges from $757/2424/2576$ to $301/76/228$ after finishing each layer ($399/1240/1316$ to $305/118/194$ for vertices).
While the described condition is sufficient for pruning, it may perhaps be improved further, providing an alternative to Analytical Marching \cite{Lei2020}.

\def\width{.6\columnwidth}
\begin{figure}
\begin{center}
  \includegraphics[width=\width]{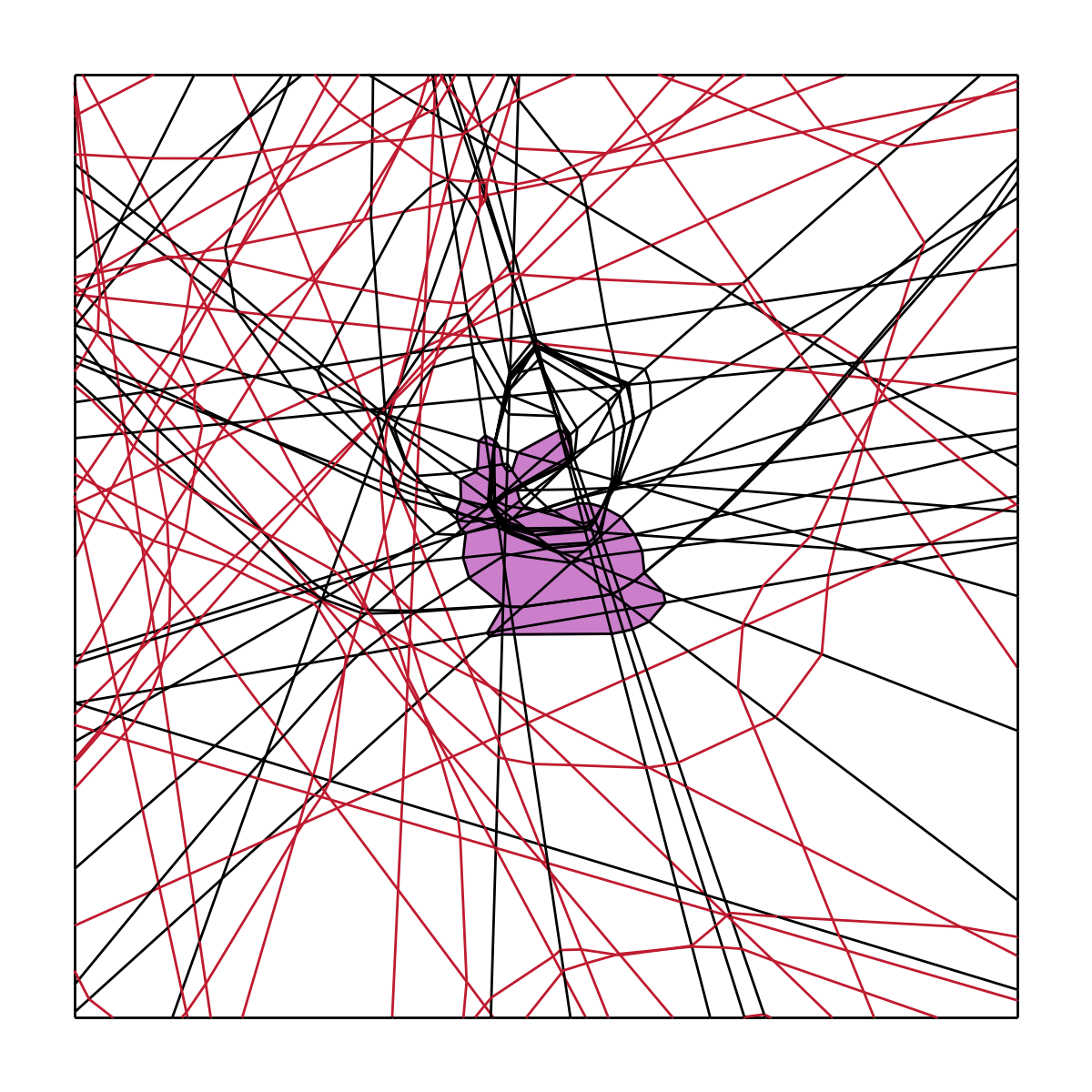}
\end{center}
\caption{
Parameter pruning. Highlighted in red are folded hyperplanes for which the purple shape is contained fully in the hyperplane's negative halfspace.
Since after \gls{ReLU} such a neuron is 0 everywhere, it can simply be removed.
A similar pruning approach can be taken to non-intersecting positive neurons.
}
\label{fig:pruning}
\end{figure}

\def\width{1.0\columnwidth}
\begin{figure}
\begin{center}
  \includegraphics[width=\width]{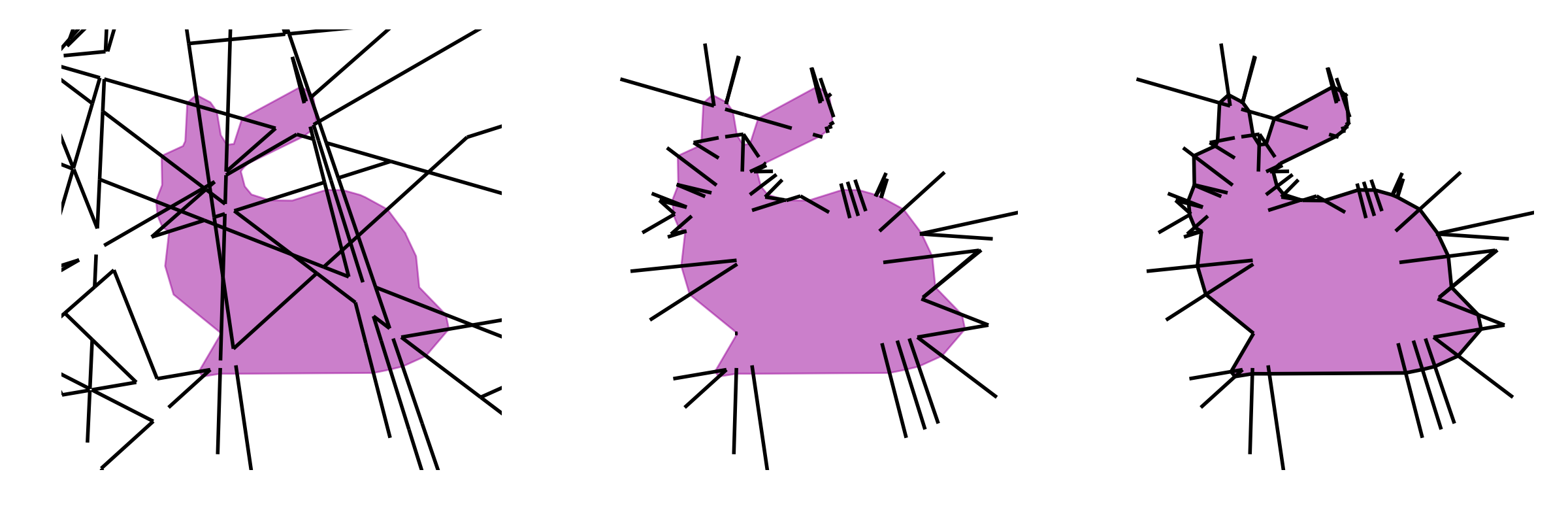}
\end{center}
\caption{Pruning during edge subdivision reduces the number of vertices and edges in each iteration by looking at future sign-vectors. Displayed are the preserved intermediate edges after each layer.}
\label{fig:pruningdivision}
\end{figure}

\section{Conclusions}
\label{sec:conclusion}

In this work, we observed a redundancy in region subdivision and proposed a novel edge subdivision method for extracting the exact polyhedral complex from \gls{ReLU} \glspl{NN}.
Our approach allowed to use simple data structures and tensor operations to leverage the GPU improving the performance over previous methods over 20 times.
The speed and differentiability allowed us to propose novel applications in which a loss can be formulated on the extracted complex.
While we hope this opens interesting avenues in geometry, in higher dimensions the method is limited by the exponential growth of the complex.
Further limitations and future research directions include extending the method to unbounded domains, non-generic arrangements, and other \gls{CPWA} architectures, as well as improving the pruning strategies for more efficient extraction of level-sets.
However, the main outlook for improved performance is replacing sorting with hash-tables, improving the whole implementation to linear time and memory in the number of vertices.

\section*{Acknowledgements}
This was supported by the European Union’s Horizon 2020 Research and Innovation Programme under Grant Agreement number 860843.

\bibliographystyle{icml2023}
\bibliography{references}

\newpage
\appendix
\section{Complexity analysis}
\label{app:complexity}

The algorithm operates on the vertices and (splitting) edges, so it is natural to first consider the complexity w.r.t. these inputs.
Let $|\mathcal{V}|$, $|\mathcal{E}|$ and $|\hat{\mathcal{E}}|$ denote the number of vertices, edges and splitting edges at iteration $i$. The complexities of the steps are:
\begin{itemize}
    \item [(1)] $O(|\mathcal{V}|)$ to evaluate the NN at $|\mathcal{V}|$ vertices (requires matrix multiplications and non-linearities).
However, complexity of each evaluation also depends on the size of the NN, but let us assume this is much less than $|\mathcal{V}|$.
    \item [(2)] $O(|\mathcal{V}|)$ to get signs of $|\mathcal{V}|$ values
    \item [(3)] $O(|\mathcal{E}|)$ to compare two signs per edge to identify splitting edges
    \item [(4)] $|\hat{\mathcal{E}}|$ to linearly interpolate $|\hat{\mathcal{E}}|$ vertex pairs of the $|\hat{\mathcal{E}}|$ splitting edges
    \item [(5)] $O((D-1)|\hat{\mathcal{E}}|)$ to build the intersecting edges.
As described, there are at most $2(D-1)|\hat{\mathcal{E}}|$ 2-faces after perturbation.
Each 2-face is associated with exactly 2 splitting edges, which we need to pair up. Using hash tables, this can be performed in linear time.
\end{itemize}

It is non-trivial to relate $|\mathcal{E}|$, $|\mathcal{V}|$ and $|\hat{\mathcal{E}}|$, but we will assume that there is a linear relationship.
We see this empirically for $|\mathcal{E}|$ and $|\mathcal{V}|$ in Figure \ref{fig:counts}.
The number of splitting edges can be upper bounded by $|\hat{\mathcal{E}}| < |\mathcal{E}| D / i$ using Theorem 5 in \cite{Hanin2019DeepReLU} where $D$ is the input dimension and $i>>D$ is the iteration (i.e., number of neurons considered already). This upper bound agrees with some empirical tests.
Hence, we replace $O(|\mathcal{E}|)$ and $|\hat{\mathcal{E}}|$ with $O(|\mathcal{V}|)$.
Since all the steps in the algorithm are linear, the algorithm in total is linear $O(|\mathcal{V}|)$.

\end{document}